\documentclass[10pt, conference]{IEEEtran}
\IEEEoverridecommandlockouts
\usepackage{cite}
\usepackage{amsmath,amssymb,amsfonts}
\usepackage{algorithmic}
\usepackage{graphicx}
\usepackage{textcomp}
\usepackage{xcolor}
\usepackage{listings}
\usepackage{multirow}
\usepackage{array}
\usepackage{ragged2e}
\usepackage{hyperref}
\usepackage{tablefootnote}
\def\BibTeX{{\rm B\kern-.05em{\sc i\kern-.025em b}\kern-.08em
    T\kern-.1667em\lower.7ex\hbox{E}\kern-.125emX}}
\begin{document}
\definecolor{lightgray}{rgb}{0.87, 0.87, 0.87}
\title{The Heap: A Contamination-Free Multilingual Code Dataset for Evaluating Large Language Models
}

\author{\IEEEauthorblockN{Jonathan Katzy}
\IEEEauthorblockA{\textit{Delft University of Technology}\\
Delft, The Netherlands \\
0009-0005-9574-2414}
\and
\IEEEauthorblockN{Razvan Mihai Popescu}
\IEEEauthorblockA{\textit{Delft University of Technology}\\
Delft, The Netherlands \\
0009-0003-6251-770X}
\and
\IEEEauthorblockN{Arie van Deursen}
\IEEEauthorblockA{\textit{Delft University of Technology}\\
Delft, The Netherlands \\
0000-0003-4850-3312}
\and
\IEEEauthorblockN{Maliheh Izadi}
\IEEEauthorblockA{\textit{Delft University of Technology}\\
Delft, The Netherlands \\
0000-0001-5093-5523}
}
\maketitle

\begin{abstract}
The recent rise in the popularity of large language models has spurred the development of extensive code datasets needed to train them. This has left limited code available for collection and use in the downstream investigation of specific behaviors, or evaluation of large language models without suffering from data contamination. To address this problem, we release \textit{The Heap}, a large multilingual dataset covering $57$ programming languages that has been deduplicated with respect to other open datasets of code, enabling researchers to conduct fair evaluations of large language models without significant data cleaning overhead.
\end{abstract}


\section{Introduction}
The data-intensive training process of Large Language Models (LLMs) has driven the release of numerous large-scale datasets, particularly for code, to facilitate the development of new models. 
This rapid increase in the amount of training data used to pre-train LLMs has resulted in extensive datasets covering almost all publicly available code~\cite{thestack, together2023redpajama, lozhkov2024starcoder2}. 

To fairly evaluate LLMs in downstream tasks, \emph{fresh} data not seen during training is needed.
Otherwise such evaluations are \emph{contaminated}, possibly resulting in overly optimistic results.
Unfortunately, obtaining non-contaminated data is increasingly difficult.
In fact, a recent study establishes that only 10\% of investigations involving LLMs deduplicate their data with respect to the training data in order to avoid contamination~\cite{vitale2024catalog}.

To address this, we propose \emph{The Heap}, a dataset of unseen code that can be used for contamination-free evaluation of LLMs in downstream tasks.
We address contamination in two ways.
First, we select code with a \emph{non-permissive} license, such as the GNU General Public License. Using such code for \emph{training} is unattractive, as it may require the end user to publicly release \emph{all} code in their code bases.
Second, we pre-conduct computationally expensive near and exact \emph{deduplication}, marking code that is used in other datasets widely used for training such \mbox{as The Stack~\cite{thestack}.}

\section{Collection}
Using the search API, we collect code from GitHub, an online platform for sharing code repositories. This process mimics the collection of other large-scale datasets~\cite{kocetkov2023the, lozhkov2024starcoder2, gao2020pile, codeparrot, ghcode}, minimizing the probability of including confounding factors in the dataset, such as drifts in the representations of data~\cite{bolukbasi2021illusion}.

\subsection{Programming Languages}
We aim to compile a representative dataset that encompasses a wide range of programming languages. To achieve this, we select languages based on several criteria. Our selection includes languages with diverse syntactic structures, such as LISP, C, Python, Haskell, and Assembly. We also select different programming paradigms, such as COBOL, Pascal, and C for procedural languages, Java, C\#, Python, for object-oriented languages, and Haskell and Clojure for functional languages. To cover more specific use cases, we also include domain-specific languages such as Mathematica, Emacs-Lisp, and Coq.
A complete list of all languages included in the dataset is presented in Table~\ref{tab:languages}.

\begin{table}[ht]
    \centering
    \caption{Languages included in the dataset}

    \begin{tabular}{l|r|r|r}
        \textbf{Language} & \textbf{Repositories} & \textbf{Raw Files} & \textbf{Unique Files} \\
        \hline
        Ada & 676 & 41,367 & 34,068\\
        Agda & 142 & 5,483 & 3,021 \\
        ANTLR & 101 & 564 & 538 \\
        Apex & 253 & 17,833 & 7,561 \\
        Assembly & 7,100 & 208,896 & 101,093 \\
        C & 50,000 & 16,585,280 & 3,076,470 \\
        C\# & 50,000 & 5,906,716 & 3,257,456 \\
        C++ & 50,000 & 14,891,856 & 4,469,823 \\
        Clojure & 27,107 & 380,567 & 269,118 \\
        Cobol & 341 & 2,242 & 1,172 \\
        Common Lisp & 796 & 45,083 & 13,922 \\
        Coq & 477 & 54,137 & 22,549 \\
        Crystal & 368 & 11,606 & 6,818 \\
        Cuda & 1,191 & 26,948 & 12,418 \\
        D & 1,185 & 185,630 & 54,034 \\
        Dart & 11,907 & 484,935 & 412,675 \\
        EJS & 1,475 & 15,513 & 12,832 \\
        Elixir & 2,371 & 643,856 & 102,874 \\
        Emacs Lisp & 377 & 8,260 & 7,312 \\
        Erlang & 1,240 & 55,932 & 27,322 \\
        F\# & 876 & 22,152 & 13,282 \\
        Forth & 222 & 28,287 & 5,129 \\
        Go & 50,000 & 8,506,379 & 2,328,529 \\
        Groovy & 2,198 & 60,299 & 47,366 \\
        Hack & 1,379 & 84,916 & 37,189 \\
        Haskell & 8,023 & 122,788 & 106,583 \\
        Java & 50,000 & 6,989,601 & 5,168,193 \\
        JavaScript & 50,000 & 8,289,901 & 1,907,803 \\
        Julia & 2,859 & 46,284 & 36,830 \\
        Kotlin & 21,665 & 1,467,343 & 1,042,136 \\
        Less & 433 & 17,276 & 7,308 \\
        Lua & 42,241 & 4,605,230 & 905,120 \\
        Mathematica & 1,528 & 164,498 & 21,208 \\
        MATLAB & 20,828 & 1,051,354 & 599,085 \\
        NetLogo & 332 & 900 & 855 \\
        NewLisp & 35 & 5,819 & 5,123 \\
        Nix & 1,892 & 75,093 & 70,407 \\
        Objective-C & 7,700 & 1,899,714 & 520,332 \\
        OCaml & 1,961 & 121,890 & 60,863 \\
        Pascal & 5,218 & 330,832 & 180,652 \\
        Perl & 14,673 & 1,798,520 & 224,753 \\
        PHP & 50,000 & 12,707,727 & 3,310,243 \\
        Processing & 2,950 & 24,723 & 20,304 \\
        Prolog & 1,071 & 38,995 & 17,570 \\
        Python & 50,000 & 2,290,182 & 1,595,919 \\
        R & 44,993 & 589,139 & 11,679 \\
        Raku & 158 & 1,384 & 689 \\
        Ruby & 13,378 & 1,579,655 & 662,915 \\
        Rust & 42,847 & 2,496,177 & 802,707 \\
        Scala & 5,893 & 749,370 & 210,630 \\
        Scheme & 1,878 & 106,620 & 50,222 \\
        Scilab & 199 & 4,531 & 3,896 \\
        SQL & 130 & 47,185 & 40,800 \\
        Starlark & 146 & 524 & 487 \\
        Swift & 13,924 & 633,819 & 434,849 \\
        Vue & 14,858 & 457,605 & 321,502 \\
        WebAssembly & 68 & 834 & 544\\
        \hline 
        \textbf{Total} & 733,663 & 96,990,250 & 32,666,778 \\
    \end{tabular}
    \label{tab:languages}
\end{table}

\subsection{Query}
We focus on repositories that have one of the targeted languages as the main language of the repository. We further select only repositories that are licensed under non-permissive licenses. We choose non-permissive licenses as an initial filter for repositories, as many large-scale datasets focus on exclusively unlicensed or permissively \mbox{licensed code~\cite{kocetkov2023the, lozhkov2024starcoder2, together2023redpajama}.} Non-permissively licensed code is usually removed due to potential licensing issues that may be related to the output of models trained on non-permissively licensed data~\cite{ourPaper}. \textit{The Heap} is not intended for pre-training models that are aimed at end users, but rather for exclusive use in a research setting. The inclusion of exclusively non-permissively licensed code has the added benefit that it acts as a deterrent for developers to train LLMs on \textit{The Heap}, ensuring it remains a relevant source of data for downstream tasks.
We provide an overview of the licenses used in this work in Table~\ref{tab:licenses}.

\subsection{Scraping}
For each programming language, we scrape up to 50,000 repositories or as many as are available. 
Our dataset contains code from repositories created between January 2008 and August 2024. For each selected language, we extract repositories sorted by star count in descending order; this has been used as a loose quality metric before~\cite{li2023starcoder}. To maximize extraction efficiency and avoid GitHub’s rate limits, we employ pagination and repository creation date filtering. When the number of repositories within a specified time frame and star range exceeds the rate limit, we narrow the time interval and apply a tumbling window approach to ensure comprehensive coverage. We guide the file extraction based on a list of file extensions from The Stack~\cite{kocetkov2023the}.

\lstset{
  basicstyle=\ttfamily\footnotesize,
  numberstyle=\tiny\color{gray},
  keywordstyle=\bfseries\color{blue},
  backgroundcolor=\color{lightgray},
  showstringspaces=false,
  breaklines=true,
  frame=single,
  rulecolor=\color{black},
  tabsize=2,
  captionpos=b,
  %doi
  morekeywords=[16]{id, full_name, html_url, stargazers_count, forks_count, 
                watchers_count, open_issues_count, language, created_at, pushed_at, 
                license, key, name, spdx_id, url, node_id, retrieval_date, file_name, file_path, content, size, extension, total_lines, avg_line_length, max_line_length, alphanum_fraction, repo_name, repo_stars, repo_forks, repo_open_issues, repo_license, repo_extraction_date, exact_duplicates_stackv2, exact_duplicates_stackv1, near_duplicates_stackv2, near_duplicates_stackv1},
  string=[s]{"}{"}, 
  stringstyle=\color{black},
  numbers=left,
  stepnumber=1,
  commentstyle=\bfseries\color{black},
  morecomment=[l]{//},
}

\subsection{Cleaning}
After collecting the data from online sources, we perform some cleaning steps. First, we exclude files containing fewer than 10 words or exceeding $10$ MB in size. We also remove exact duplicates from our own dataset. We use the same approach as the exact deduplication with respect to other datasets described in Section~\ref{sec:datasets}.

\begin{table}[]
    % \centering
    \caption{Copyleft licenses included in the dataset.}
    \begin{tabular}{>{\raggedright}m{1.5cm}|>{\raggedright}m{2.05cm}|>{\raggedright\arraybackslash}m{3.8cm}}
     \textbf{License} & \textbf{Family} & \textbf{Description}\tablefootnote{\url{https://blueoakcouncil.org/copyleft}} \\
      \hline
      CECILL-1.0 CECILL-1.1 CECILL-2.0 CECILL-2.1 CECILL-C EPL-1.0 EPL-2.0 LGPL-2.1 LGPL-3.0 MS-RL MPL-2.0 & Weak Copyleft & Share changes and additions to the licensed software when redistributing. \\
      \hline
      GPL-2.0 GPL-3.0 & Strong Copyleft & Share larger programs built with the licensed software when redistributing. This extends weak copyleft requirements. \\ 
      \hline
      AGPL-3.0 EUPL-1.1 EUPL-1.2 OSL-3.0 & Network Copyleft & Share larger programs built with the licensed software when redistributing or running it over a network. This extends strong copyleft requirements. \\
    \end{tabular}
    \label{tab:licenses}
\end{table}

\section{Deduplication}
An important aspect of fairly evaluating downstream tasks is preventing data leakage~\cite{vitale2024catalog}. This is often done through a deduplication process. Although there should be no overlap between our non-permissively licensed dataset and permissively licensed datasets due to our selection procedure, it does not completely prevent overlap~\cite{ourPaper}.

Our deduplication strategy consists of exact deduplication and near deduplication. Before each deduplication strategy, we remove all comments (using a regex, based on the programming language) and whitespace from each file. This ensures that small changes to files, such as the removal of a license comment or changes in whitespace characters, still result in the detection of an exact duplicate. The final files included in \textit{The Heap} are the unaltered versions scraped from GitHub.

\paragraph{Exact Deduplication}
For exact deduplication, we calculate the SHA-256 hash of each file to identify exact duplicates between \textit{The Heap} and publicly available datasets. We selected this hash function for its low collision probability, which reduces the risk of false positives.

\paragraph{Near Deduplication}
We also perform near-deduplication between our scraped dataset and the publicly available ones. To achieve this, we utilize the MinHash Locality-Sensitive Hashing (LSH) approach, implemented using the \texttt{datasketch}\footnote{\url{https://ekzhu.com/datasketch/lsh.html}} library. We apply the same SHA-256 hashing function as before, with $128$ permutations and a precision-recall weight distribution of $40\%-60\%$. These design choices help mitigate hash collisions while maintaining a balanced trade-off, hence favoring higher recall at the expense of a controlled increase in false positives (removing files that were not duplicates). 

We use a shingle size of $7$ characters, as code files typically use a smaller set of characters compared to natural language, 
where $k = 9$~\cite{massiveDatasets}. This reduces the likelihood of overly common shingles, which could otherwise inflate similarity scores, as would occur with smaller values of $k$. Files with a Jaccard similarity above $0.7$ are flagged as near duplicates, a threshold shown to be effective for duplicate detection~\cite{allamanis}. 

\begin{figure}
\begin{lstlisting}
 { 
     id: 200,
     file_name: "kernel.lisp",
     file_path: "whily_yalo/cc/kernel.lisp",
     content: "REPL: loop (jmp short read-start) ;; ...",
     size: 4,099,
     language: "Common Lisp",
     extension: ".lisp",
     total_lines: 125, 
     avg_line_length: 27.52,
     max_line_length: 104,
     alphanum_fraction: 0.59,
     repo_name: "whily/yalo",
     repo_stars: 571,
     repo_forks: 32, 
     repo_open_issues:1,
     repo_license: "GPL-2.0",
     repo_extraction_date: "9/19/2024, 11:24:32 AM",
     exact_duplicates_stackv1: False,
     exact_duplicates_stackv2: True,
     near_duplicates_stackv1: False,
     near_duplicates_stackv2: True,
     ... 
   }
\end{lstlisting}
\caption{Example of final dataset structure for one entry}
\label{lst:final_structure}
\end{figure}

\begin{table}[]
    \centering
    \caption{List of publicly-available datasets used for deduplication}
    \begin{tabular}{l|p{.6\hsize}}
        \textbf{Dataset} & \textbf{Source} \\
        \hline
        The Stack V2~\cite{lozhkov2024starcoder2} & All permissively licensed and unlicensed files collected in the \mbox{Software Heritage~\cite{softheritage}} archive. \vspace{2.5pt}\\
        % CodeClippy & 16 TB \\
        The Stack~\cite{thestack}& All permissively licensed repositories collected in the GHArchive~\cite{gharchive} and scraped from GitHub. \vspace{2.5pt}\\
        Red Pajama~\cite{together2023redpajama}& Repositories from the GitHub dataset hosted by Google BigQuery~\cite{bigquery} licensed under MIT, BSD, or Apache licenses. \vspace{2.5pt}\\ 
        GitHub Code~\cite{ghcode }& Repositories from the GitHub dataset hosted by Google BigQuery~\cite{bigquery}. \vspace{2.5pt}\\
        % The Pile & 825 GB\\ 
        CodeParrot~\cite{codeparrot} & All Python files from the GitHub dataset hosted by Google BigQuery~\cite{bigquery}.\\ 
    \end{tabular}
    \label{tab:pub_datasets}
\end{table}

We identify and flag duplicates between our dataset and all publicly available datasets to facilitate a more flexible approach to LLM evaluation, prioritizing both reproducibility and ease of use. This setup minimizes time and computational overhead by removing the burden of duplicate detection from researchers. Users can seamlessly filter data by language or by exact and near-duplicate files, tailoring the dataset to their specific requirements. Table \ref{tab:languages} provides a comprehensive summary of the languages extracted. The third column lists the number of files collected after filtering based on file size and word count. The last column indicates the number of files obtained after removing exact duplicates within our dataset, with exact and near duplicates from other datasets flagged among the remaining files.
For more detailed information on the dataset creation process, please refer to the dataset page\footnote{\url{https://huggingface.co/datasets/AISE-TUDelft/the-heap}}.

\subsection{Datasets}\label{sec:datasets}
Our selection of datasets for deduplication is based on previously curated lists~\cite{ourPaper}, with the addition of \mbox{The Stack V2~\cite{lozhkov2024starcoder2},} which is the only new dataset that has been released since the publication of previous works. We give an overview of all potential datasets in Table~\ref{tab:pub_datasets}. Due to the comment removal being based on the programming languages of the files, we are not able to infer the correct language for two datasets. The Pile~\cite{gao2020pile}, which has been removed and re-uploaded, has lost information about the programming language of a file.
Furthermore, due to a known issue with the curation of CodeClippy\footnote{\url{https://github.com/CodedotAl/gpt-code-clippy/issues/71}}, the languages and names of files are misaligned in the dataset. We also exclude this dataset from deduplication. Although we could predict the languages used in the files in these datasets, the tools that provide this functionality do return incorrect predictions, which could result in a duplicate not being removed. As we aim to provide a guarantee that there is no data contamination in our dataset, we remove these two datasets from consideration.

\section{Layout}
\textit{The Heap} is organized into multiple subsets, each of them corresponding to one programming language. In each subset, the entries included in the dataset can be summarized into 3 groups: file content and metadata, quality indicators, and duplicates. We give an example of one entry in Figure~\ref{lst:final_structure}.

\paragraph{File Content and Metadata}
For the file content and metadata, we list the actual content of the file, which is the main information to be used in downstream tasks. We also include information about filename and path, as this has been included in the pre-training procedure of \mbox{some LLMs~\cite{li2023starcoder, team2024codegemma, lozhkov2024starcoder2}.}

\paragraph{Quality Indicators}
To facilitate the selection of files for downstream use, we incorporated several quality indicators previously utilized in related works, ensuring the dataset can be easily filtered and selected.
We included numerical statistics about the file such as the \textit{total\_lines}, \textit{avg\_line\_length}, \textit{max\_line\_length} and \textit{alphanumeric fraction}, as well as repository-wide statistics such as \textit{repo stars}, \textit{repo forks}, \textit{open issues} and the \textit{extraction date of the repo}. The repository star count will be artificially inflated for languages where more than $50,000$ repositories exist, due to the ordering of the repositories in the collection steps.

\paragraph{Duplicates}
As we deduplicate \textit{The Heap} with respect to a number of other publicly available datasets, we incorporate two columns for every dataset. One column contains a Boolean value, whether there is an exact duplicate of the given file in the dataset, and the other column contains a Boolean value describing whether there is a near duplicate of the given file in the dataset. We choose not to remove files but to use a Boolean mask in order to maximize the amount of available data for each available dataset.

\section{Future Improvements}
In future iterations of this dataset, several potential improvements could be made. These include enhancing the deduplication process, releases of new training datasets, providing detailed information about the natural languages represented in the dataset, and tracking the evolution of codebases.

\paragraph{New Datasets}
The main goal of this dataset is to reduce the burden of deduplicating a dataset used for downstream tasks for future research. This is only effective if the dataset is deduplicated against all available datasets. As new datasets are released we intend to pass them through the same pipeline to ensure \textit{The Heap} remains relevant for the future.

\paragraph{Deduplication}
We addressed the deduplication of datasets using two widely adopted methods: exact deduplication based on hashing and near-deduplication leveraging locality-sensitive hashing. However, there is limited research on what constitutes an effective deduplication strategy. There could be issues with duplicates at a lower granularity level than file-based deduplications, as well as possible issues with the provenance of code fragments. Once studies are conducted on the impact of various deduplication approaches, we plan to incorporate these strategies as a new entry in the dataset.

\paragraph{Cleaning}
We include all files that we scraped that were not duplicates, while this gives us a dataset of deduplicated files, there is still the question of file ``quality''. In NLP research, keywords have been used for filtering websites, such as \texttt{lorem ipsum} or \texttt{TODOs}~\cite{dodge2021documenting}, and code datasets have been cleaned of autogenerated files using a similar approach~\cite{lozhkov2024starcoder2}. We believe that this may also affect the quality of code datasets. Specifically, languages that rely heavily on boiler plating, such as Java, may benefit from removing certain common phrases from their corpus. This will be included as a further filtering step in a future release of the dataset.

\paragraph{Topic Modeling}
While languages can be used to loosely select an area that is being analyzed (Mathematica for mathematics, or JavaScript for web-based projects), many languages can be used in multiple specializations/areas. Adopting the FineWeb topic modeling approach for code datasets would create interesting annotations for the code files, as well as show any form of topic-based imbalances in the dataset.

\paragraph{Natural Language}
An under-explored research area involves the presence of multiple natural languages within code. As natural languages are often mixed within one file~\cite{pawelka2015code}, we plan to adopt a Parts of Speech-like tagging~\cite{chiche2022part} system for the natural languages present in each file. This can give information about the performance of code LLMs when the code is not in English. This will both help the development of non-English code LLMs, as well as aid English-focused LLMs, as they can be evaluated on only English.

\section{Limitations/Challenges}
The limitations and challenges faced by this dataset are two-fold. First, other actors may decide to train their models on this data, removing the benefits, and second, developers may object to their code being present in this dataset. We address these problems as follows.

\paragraph{Training}
In order to use \textit{The Heap} for a fair evaluation of an LLM, the researcher must be sure that the target LLM has not been trained on \textit{The Heap}. Aside from our deduplication ensuring this fact for current existing LLMs, our collection process also adds a layer of protection from the inclusion of \textit{The Heap} in the training procedure. The trend of training LLMs has shifted to only training on permissively licensed data, which would exclude \textit{The Heap}. Furthermore, the restriction of \textit{The Heap} to research only, alleviates the problems with author attribution in LLM generations as trained models are not intended to be used by end-users~\cite{longpre2023data, ourPaper}.

Furthermore, existing works such as membership inference attacks, have been extended to the scale of datasets~\cite{maini2024llm}. This should make it possible in the near future to test for the inclusion of \textit{The Heap} in the training procedures of a model.

\paragraph{Ethics}
With the rapid rise of public repositories being used to train code language models, many authors of older repositories were unaware that their code could be utilized for such purposes, leaving them unable to opt-out. Moreover, there is currently no consensus on how developers can opt in or out of having their code included in datasets. We acknowledge these ethical concerns regarding the use of code in deep learning practices and offer the ability for repository owners to opt out of having their code included in our dataset. Although this approach is not ideal, as it places the burden of exclusion on the authors, it aligns with the current best practices~\cite{lozhkov2024starcoder2}.

\section{Conclusion}
We present \textit{The Heap}, a multilingual dataset of source code that we deduplicated against datasets commonly used in the (pre-)training of large language models. \textit{The Heap} enables researchers to conduct investigations into the behavior and performance of code large language models without the need to perform extensive deduplication with other datasets. This addresses the shortcomings of LLM investigations not testing for data leakage in 90\% of all investigations\cite{vitale2024catalog} allowing for more robust conclusions to be made.

We release the dataset (only for research purposes) and outline a road map for future features such as natural language annotation, topic annotations, and further cleaning procedures to be incorporated into the dataset, to make higher-quality evaluations easier and more available for all researchers. 

\bibliography{main}
\bibliographystyle{unsrt}
\end{document}